\documentclass[10pt,twocolumn,letterpaper]{article}

\usepackage[pagenumbers]{cvpr}
\usepackage{times}
\usepackage{epsfig}
\usepackage{graphicx}
\usepackage{amsmath}
\usepackage{amssymb}
\usepackage{url}
\usepackage{booktabs}
\usepackage{amsfonts}
\usepackage{nicefrac}
\usepackage{microtype}
\usepackage{comment}
\usepackage{multirow}
\usepackage{wrapfig}
\usepackage{float}
\usepackage{multicol}
\usepackage{dblfloatfix}
\usepackage[font=small,labelfont=bf]{caption}
\usepackage[page]{appendix}

\usepackage[dvipsnames]{xcolor}

\definecolor{cvprblue}{rgb}{0.21,0.49,0.74}
\usepackage[pagebackref,breaklinks,colorlinks,citecolor=cvprblue]{hyperref}

\newcommand{\OURS}{CG-HOI}

\title{\OURS: Contact-Guided 3D Human-Object Interaction Generation}

\author{Christian Diller\\
Technical University of Munich\\
{\tt\small christian.diller@tum.de}
\and
Angela Dai\\
Technical University of Munich\\
{\tt\small angela.dai@tum.de}
}

\begin{document}

\twocolumn[{%
	\renewcommand\twocolumn[1][]{#1}%
	\maketitle
	\begin{center}
    \includegraphics[width=\linewidth]{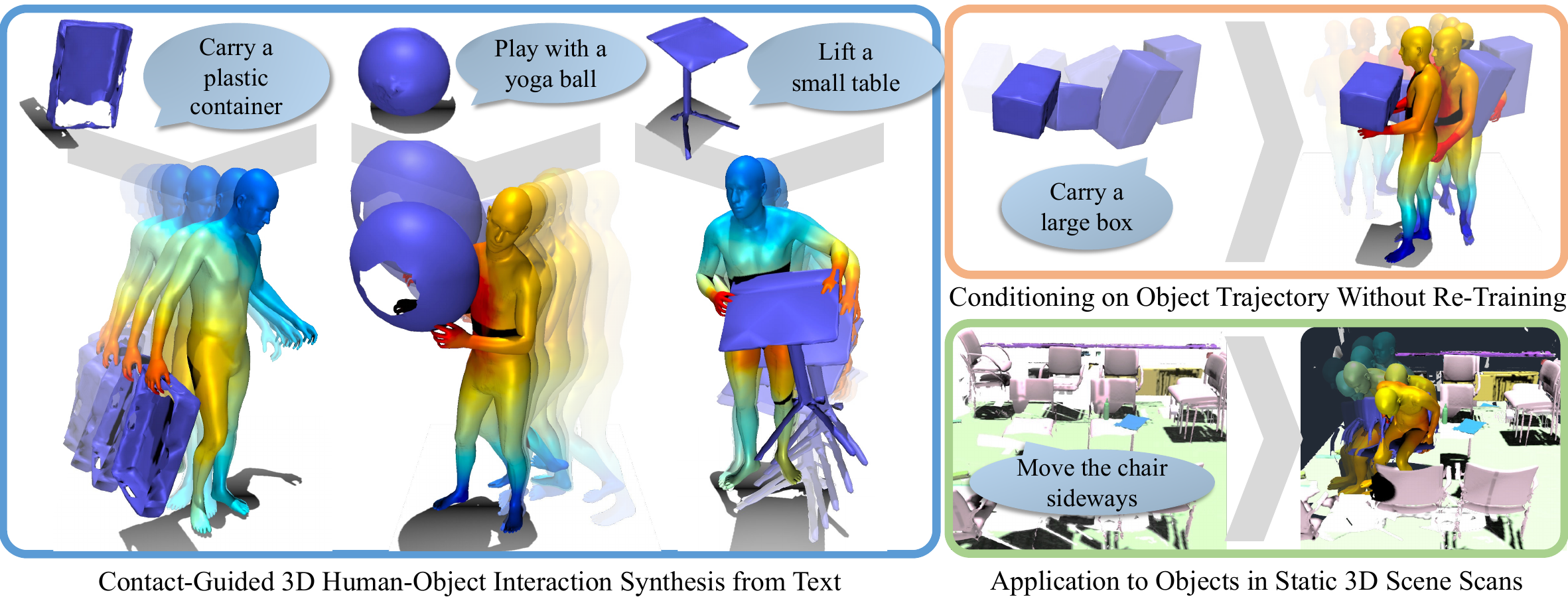}
    \vspace{-1.5em}
    \captionof{figure}{We present an approach to generate realistic 3D human-object interactions (HOIs), from a text description and given static object geometry to be interacted with (left). Our main insight is to explicitly model contact (visualized as colors on the body mesh, closer contact in red), in tandem with human and object sequences, in a joint diffusion process. In addition to synthesizing HOIs from text, we can also synthesize human motions conditioned on given object trajectories (top right), and generate interactions in static scene scans (bottom right).}
    \label{fig:title}
	\end{center}
}]

\begin{abstract}
\vspace{-1em}
    We propose \OURS{}, the first method to address the task of generating dynamic 3D human-object interactions (HOIs)  from text. 
    We model the motion of both human and object in an interdependent fashion, as semantically rich human motion rarely happens in isolation without any interactions. 
    Our key insight is that explicitly modeling contact between the human body surface and object geometry can be used as strong proxy guidance, both during training and inference.
    Using this guidance to bridge human and object motion enables generating more realistic and physically plausible interaction sequences, where the human body and corresponding object move in a coherent manner.
    Our method first learns to model human motion, object motion, and contact in a joint diffusion process, inter-correlated through cross-attention. 
    We then leverage this learned contact for guidance during inference to synthesize realistic and coherent HOIs.
    Extensive evaluation shows that our joint contact-based human-object interaction approach generates realistic and physically plausible sequences, and we show two applications highlighting the capabilities of our method. Conditioned on a given object trajectory, we can generate the corresponding human motion without re-training, demonstrating strong human-object interdependency learning. Our approach is also flexible, and can be applied to static real-world 3D scene scans.
\vspace{-1em}
\end{abstract}

\section{Introduction}
\label{sec:intro}

Generating human motion sequences in 3D is important for many real-world applications, e.g. efficient realistic character animation, assistive robotic systems, room layout planning, or human behavior simulation. 
Crucially, human interaction is interdependent with the object(s) being interacted with; the object structure of a chair or ball, for instance, constrains the possible human motions with the object (e.g., sitting, lifting), and the human action often impacts the object motion (e.g., sitting on a swivel chair, carrying a backpack).

Existing works typically focus solely on generating dynamic humans, and thereby disregarding their surroundings \cite{zhang2023tedi,raab2023single,zhao2023modiff,dabral2023mofusion,ren2023diffusion,chen2023executing}, or grounding such motion generations in a static environment that remains unchanged throughout the entire sequence \cite{huang2023diffusion,xiao2023unified,wang2022towards,wang2022humanise,zhao2022compositional,zhao2023synthesizing,zhang2022wanderings,mullen2023placing,zheng2022gimo}.
However, real-world human interactions affect the environment. For instance, even when simply sitting down on a chair, the chair is typically moved: to adjust it to the needs of the interacting human, or to move it away from other objects such as a table. 
Thus, for realistic modeling of human-object interactions, we must consider the interdependency of object and human motions.

We present \OURS{}, the first approach to address the task of generating realistic 3D human-object interactions from text descriptions, by jointly predicting a sequence of 3D human body motion along with the object motion.
Key to our approach is to not only model human and object motion, but to also explicitly model contact as a bridge between human and object.
In particular, we model contact by predicting contact distances from the human body surface to the closest point on the surface of the object being interacted with. 
This explicit modeling of contact helps to encourage human and object motion to be semantically coherent, as well as to provide a constraint indicating physical plausibility (e.g., discouraging objects to float without support).

\OURS{} jointly models human, object, and contact together in a denoising diffusion process. Our joint diffusion model is designed to encourage information exchange between all three modalities through cross-attention blocks. 
Additionally, we employ a contact weighting scheme, based on the insight that object motion, when being manipulated by a human, is most defined by the motion of the body part in closest contact (Fig.~\ref{fig:contact-weighting}). We make use of this by generating separate object motion hypotheses for multiple parts of the human body and aggregating them based on that part's predicted contact.
During inference, we leverage the predicted contact distances to refine synthesized sequences through our contact-based diffusion guidance, which penalizes synthesizing sequences with human-object contact far from the predicted contact distances.

Our method is able to generate realistic and physically plausible human-object interactions, and we evaluate our approach on two widely-used interaction datasets, BEHAVE~\cite{bhatnagar2022behave} and CHAIRS~\cite{jiang2022chairs}. In addition, we also demonstrate the usefulness of our model with two related applications: First, generating human motion given a specific object trajectory without any retraining, which demonstrates our learned human-object motion interdependencies. Second, populating a static 3D scene scan with human-object interactions of segmented object instances, showing the applicability of our method to general real-world 3D scans.

In summary, our contributions are three-fold:
\begin{itemize}
    \item We propose an approach to generate realistic, diverse, and physically plausible human-object interaction sequences by jointly modeling human motion, object motion, and contact through cross-attention in a diffusion process. 
    \item We formulate a holistic contact representation: Object motion hypotheses are generated for multiple pre-defined points on the surface of the human body and aggregated based on predicted contact distances, enabling comprehensive body influence on contact while focusing on the body parts in closer contact to the object.
    \item We propose a contact-based guidance during synthesis of human-object interactions, leveraging predicted contacts to refine generated interactions, leading to more physically plausible results.
\end{itemize}

\section{Related Work}
\label{sec:related}

\begin{figure*}[t]
    \centering
    \includegraphics[width=\textwidth]{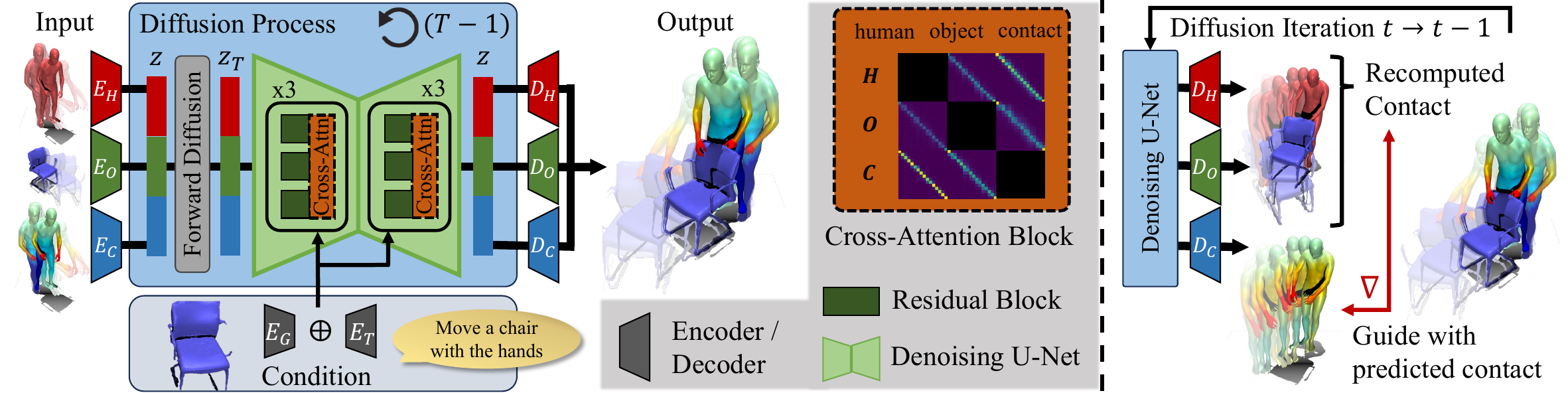}
    \vspace{-1em}
    \caption{Method Overview. 
    Given a text description and object geometry, \OURS{} produces a human-object interaction (HOI) sequence, modeling both human and object motion.
    To produce realistic HOIs, we additionally model contact to bridge the interdependent motions. Our method jointly generates all three during training (left), using a U-Net-based diffusion with cross-attention across human, object, and contact. During inference (right), we drive synthesis under guidance of estimated contact to sample more physically plausible interactions.
    }
    \label{fig:overview}
    \vspace{-1em}
\end{figure*}

\noindent
\textbf{3D Human Motion Generation.}
Generating sequences of 3D humans in motion is a task which evolved noticeably over the last few years. Traditionally, many methods used recurrent approaches \cite{fragkiadaki2015recurrent,aksan2019structured,chiu2019action,gopalakrishnan2019neural,jain2016structural,martinez2017human} and, improving both fidelity and predicted sequence length, graph- and attention-based frameworks \cite{mao2019learning,mao2020history,tang2018long}.
Notably, generation can either happen deterministically, predicting one likely future human pose sequence \cite{fragkiadaki2015recurrent,martinez2017human,mao2019learning,mao2020history,diller2024forecasting}, or stochastically, thereby also modelling the uncertainty inherent to future human motion \cite{aliakbarian2020stochastic,barsoum2018hp,mao2021generating,yan2018mt,yuan2020dlow,diller2022forecasting,bhattacharyya2018accurate,xu2022stochastic}.

Recently, denoising diffusion models \cite{sohl2015deep,song2020denoising} showed impressive results in 2D image generation, producing high fidelity and diverse images \cite{ho2020denoising,song2020denoising}. 
Diffusion models allow for guidance during inference, with classifier-free guidance \cite{ho2022classifier,nichol2021glide} widely used to trade off between generation quality and diversity.
Inspired by these advances, various methods have been proposed to model 3D human motion through diffusion, using U-Nets \cite{zhang2023tedi,raab2023single,zhao2023modiff,dabral2023mofusion,ren2023diffusion,chen2023executing}, transformers \cite{tevet2023human,tian2023transfusion,yang2023synthesizing,raab2023single,yang2023longdancediff,shafir2023human,wang2023fg,ahn2023can,zhang2022motiondiffuse,wei2023human,sun2023towards,wei2023understanding}, or custom architectures \cite{alexanderson2023listen,choi2022diffupose,chen2023humanmac,barquero2022belfusion,zhang2023remodiffuse}. Custom diffusion guidance has also been shown to aid controllability \cite{karunratanakul2023gmd,janner2022planning,rempe2023trace} and physical plausibility \cite{yuan2022physdiff}.

In addition to unconditional motion generation, conditioning on text descriptions allows for more control over the generation result \cite{tevet2023human,wang2023fg,zhang2022motiondiffuse,wei2023understanding,zhao2023modiff,ren2023diffusion}. In fact, generating plausible and corresponding motion from textual descriptions has been an area of interest well before the popularity of diffusion models \cite{guo2022generating,chen2023executing,zhang2023t2m,azadi2023make,petrovich2022temos,kim2023flame,kalakonda2022action}.

These methods show strong potential for 3D human motion generation, but focus on a skeleton representation of the human body, and only consider human motion in isolation, without naturally occurring interactions.
To generate realistic human-object interactions, we must consider the surface of the human body and its motion with respect to object motion, which we characterize as contact.

\smallskip
\noindent
\textbf{3D Human Motion in Scenes.}
As human motion typically occurs not in isolation but in the context of an object or surrounding environment, various methods have explored 
learning plausible placement of humans into scenes, both physically and semantically, \cite{zhang2020place,hassan2019resolving,zhang2020generating,hassan2021populating,hassan2023synthesizing,DBLP:journals/pacmcgit/XieTSPL23}, forecasting future motion given context \cite{cao2020long,mao2022contact}, or generating plausible walking and sitting animations \cite{huang2023diffusion,xiao2023unified,wang2022towards,wang2021synthesizing,wang2021scene,wang2022humanise,zhao2022compositional,zhao2023synthesizing,zhang2022wanderings,mullen2023placing,zheng2022gimo,hassan2021stochastic}. 
This enables more natural modeling of human reactions to their environment; however, the generated interactions remain limited due to the assumption of a static scene environment, resulting in a focus on walking or sitting movements.

Recent methods have also focused on more fine-grained interactions by generating human motion given a single static object \cite{taheri2022goal,tendulkar2023flex,zhang2022couch,wu2022saga,lee2023locomotion,zhang2023roam,kulkarni2023nifty}. 
While these methods only focus on human motion generation for a static object, \cite{DBLP:journals/tog/LiWL23} generates human motion conditioned on object motion and \cite{xu2023interdiff,DBLP:journals/ral/WanYLZJCPTKW22} generate full human-object interaction sequences directly from an initial sequence observation.
Our approach also models both human and object motion, but we formulate a flexible text-conditioned generative model for dynamic human and object motion, modeling the interdependency between human, object, and contact to synthesize more realistic interactions under various application settings.

\smallskip
\noindent
\textbf{Contact Prediction for Human-Object Interactions.}
While there is a large corpus of related work for human motion prediction, only few works focus on object motion generation \cite{driess2023learning,mrowca2018flexible,rempe2020predicting,zhu2018object}. Notably, these methods predict object movement in isolation, making interactions limited, as they typically involve interdependency with human motion.

Contact prediction has been most studied in recent years for the task of fine-grained hand-object interaction \cite{ghosh2023imos,braun2023physically,ye2023affordance,zheng2023cams,zhou2022toch,kry2006interaction,li2023task}. It is defined either as binary labels on the surface \cite{ghosh2023imos,braun2023physically,ye2023affordance,zheng2023cams,kry2006interaction,li2023task} or as the signed distance to a corresponding geometry point \cite{zhou2022toch}.
In these works, predicting object and hand states without correct contact leads to noticeable artifacts. Contact prediction itself has also been the focus of several works \cite{wu2022saga,grady2021contactopt,jiang2021hand,tse2022s}, either predicting contact areas or optimizing for them.

Applied to the task of generating whole-body human-object interactions, this requires access to the full surface geometry of both object and human. 
Only few recent motion generation works focus on generating full-body geometric representations of humans \cite{zhang2022wanderings,mao2022weakly,petrovich2021action,petrovich2022temos,tevet2022motionclip,xu2022stochastic,zhang2021we} instead of simplified skeletons which is a first step towards physically correct interaction generation.
However, while several of these works acknowledge that contact modeling would be essential for more plausible interactions \cite{petrovich2021action,petrovich2022temos,zhang2022wanderings}, they do not model full-body contact.

We approach the task of generating plausible human-object motion from only the object geometry and a textual description as a joint task and show that considering the joint behavior of full-body human, object, and contact between the two benefits output synthesis to generate realistic human-object interaction sequences.

\section{Method Overview}
\label{sec:overview}
\OURS{} jointly generates sequences of human body and object representations, alongside contact on the human body surface. Reasoning jointly about all three modalities in both training and inference enables generation of semantically meaningful human-object interaction sequences.

Fig.~\ref{fig:overview} shows a high-level overview of our approach: We consider as condition a brief text description $T$ of the action to be performed, along with the static geometry $G$ of the object to be interacted with, and generate a sequence of $F$ frames $\mathbf{x}=[\mathbf{x}_1, \mathbf{x}_2, ..., \mathbf{x}_F]$ where each frame $\textbf{x}_i$ consists of representations for the object transformation $o_i$, for the human body surface $h_i$, and for the contact $c_i$ between human and object geometry. We denote as $H=\{h_i\}$ the human body representations, $O=\{o_i\}$ the object transformations, and $C=\{c_i\}$ the contact representations.

We first train a denoising diffusion process to generate $H$, $O$, and $C$, using a U-Net architecture with per-modality residual blocks and cross-attention modules. Using cross-attention between human, object motion, and contact allows for effectively learning interdependencies and and feature sharing (Sec.~\ref{sec:diffusion}).
We use the generated contact to guide both training and inference: Instead of predicting one object motion hypothesis per sequence, we generate multiple, and aggregate them based on predicted contacts, such that body parts in closer contact with the object have a stronger correlation with the final object motion (Sec.~\ref{sec:contact-weighting}).
During inference, the trained model generates $H$, $O$, and $C$. For each step of the diffusion inference, we use predicted contact $C$ to guide the generation of $H$ and $O$, by encouraging closeness of recomputed contact and predicted contact, producing more refined and realistic interactions overall (Sec.~\ref{sec:interactiongen}).

\section{Human-Object Interaction Diffusion} 
\label{sec:diffusion}

\subsection{Probabilistic Denoising Diffusion}
Our approach uses a diffusion process to jointly generate a sequence of human poses, object transformations, and contact distances in a motion sequence from isotropic Gaussian noise in an iterative process, removing more noise at each step. More specifically, during training we add noise depending on the time step (``forward process'') and train a neural network to reverse this process, by directly predicting the clean sample from noisy input. 
Mathematically, the forward process follows a Markov chain with T steps, yielding a series of time-dependent distributions $q(\mathbf{z}_t | \mathbf{z}_{t-1})$ with noise being injected at each time step until the final distribution $\mathbf{z}_T$ is close to $\mathcal{N}(\mathbf{0}, \mathbf{I})$. Formally,
\begin{equation}
    q(\mathbf{z}_t | \mathbf{z}_{t-1}) = \mathcal{N}(\sqrt{\beta_t} \mathbf{z}_{t-1} + (1 - \beta_t)\mathbf{I})
\end{equation}
with the variance of the Gaussian noise at time $t$ denoted as $\beta_t$, and $\beta_0=0$.

Since we adopt the Denoising Diffusion Probabilistic Model \cite{ho2020denoising}, we can sample $\mathbf{z}_t$ directly from $\mathbf{z}_0$ as
\begin{equation}
    \mathbf{z}_t = \sqrt{\alpha_t}\mathbf{z}_0 + \sqrt{1-\alpha_t}\epsilon
\label{eq:forward-diffusion}
\end{equation}
with $\alpha_t=\prod_{t'=0}^t(1 - \beta_t)$, and $\epsilon \sim \mathcal{N}(\mathbf{0}, \mathbf{I})$.

For the reverse process, we follow \cite{tevet2023human,xu2023interdiff,raab2023single}, directly recovering the original signal $\mathbf{\tilde{z}}$ instead of the added noise.

\smallskip
\noindent
\textbf{Human-Object Interactions}
To model human-object interactions with diffusion, we employ our neural network formulation $\mathcal{F}$.
$\mathcal{F}$ operates on the noised vector of concatenated  human, object, and contact representations, together with the current time step $t$, and a condition consisting of object point cloud $G$, encoded by an encoder $E_G$, and text information $T$, encoded by encoder $E_T$. Formally,
\begin{equation}
    \mathbf{\tilde{z}} = \mathcal{F}(\mathbf{z}_t, t, E_G(G)\oplus E_T(T))
\end{equation}

More specifically, in our scenario $E_T$ extracts text features with a pre-trained CLIP \cite{radford2021learning} encoder. Encoder $E_G$ processes object geometry $G$ as a uniformly sampled point cloud in world coordinate space with a PointNet~\cite{pointnet2017} pre-trained on object parts segmentation.

Object transformations $o_i$ are represented as global translation and rotation using continuous 6D rotation representation \cite{zhou2019continuity}. In contrast to prior work \cite{yuan2020dlow,diller2022forecasting,tevet2023human,wang2023fg,zhang2022motiondiffuse,zhang2023roam,kulkarni2023nifty} which focused on representing human motion in a simplified manner as a collection of $J$ human joints, disregarding both identity-specific and pose-specific body shape, we model physically plausible human-object contacts between body surface and geometry. Thus, we represent the human body $h_i$ in SMPL \cite{smpl2015} parameters: $h_i=\{h_i^p, h_i^b, h_i^r, h_i^t\}$ with pose parameters $h_i^p\in \mathbb{R}^{63}$, shape parameters $h_i^b\in \mathbb{R}^{10}$, as well as global rotation $h_i^r\in \mathbb{R}^3$ and translation $h_i^t\in \mathbb{R}^3$. These body parameters can then be converted back into a valid human body surface mesh in a differentiable manner using the SMPL \cite{smpl2015} model. This allows us to reason about the contact between human body surface and object geometry.
We represent contact $c_i$ on the human body as the distance between a set of $M=128$ uniformly distributed motion markers on the body surface to the closest point of the object geometry, for each marker. Specifically, we represent contact for frame $\mathbf{x}_i$ and $j$-th contact marker ($j\in \{0..M-1\}$) $c_i^j$ as its distance from the human body surface to the closest point on the same frame's object geometry surface.

\subsection{Human-Object-Contact Cross-Attention}
\label{sec:threeway}
We jointly predict human body sequences $H=\{h_i\}$, object transformations $O=\{o_i\}$, and corresponding contact distances $C=\{c_i\}$ in our diffusion approach. 
We employ a U-Net backbone for diffusion across these outputs, with separate residual blocks for human, object, and contact representations, building modality-specific latent feature representations. 
As we aim to model the inter-dependency across human, object, and contact, we introduce custom human-object-contact cross-attention modules after every residual block where each modality attends to the other two. 

We follow the formulation of Scaled Dot-Product Attention \cite{vaswani2017attention}, computing the updated latent human body feature:

\begin{equation}
\label{eq:attention}
h_i = \text{softmax}\left(\frac{\boldsymbol{Q} \boldsymbol{K}^T}{\sqrt{d}}\right) \boldsymbol{V}, 
\end{equation}
with query $\boldsymbol{Q} = H$, and key and value $\boldsymbol{K} = \boldsymbol{V} = O \odot C$ ($\odot$ denotes concatenation), i.e. $\boldsymbol{Q}\in \mathbb{R}^{F\times d}$ and $\boldsymbol{K}, \boldsymbol{V}\in \mathbb{R}^{2F\times d}$. As in \cite{vaswani2017attention}, $d$ denotes the dimensionality of query and key. Applying this similarly to $O$ and $C$ yields the final features after each cross-attention module.

\subsection{Contact-Based Object Transform Weighting}
\label{sec:contact-weighting}
\begin{figure}
    \centering
    \includegraphics[width=\columnwidth]{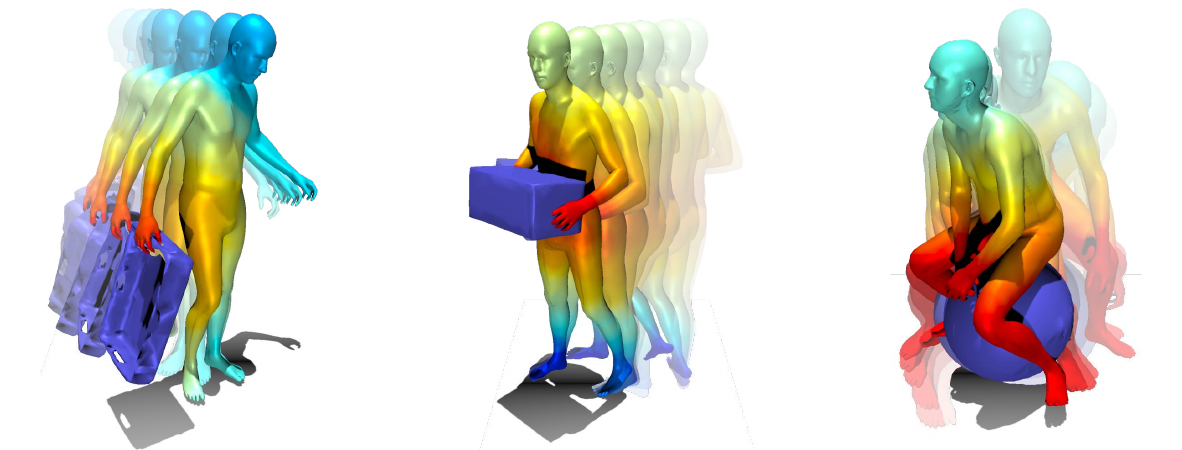}
    \vspace{-2em}
    \caption{An object's trajectory is largely defined by the motion of the region of the body in close contact with the object, e.g. the hand(s) when carrying an object (left, middle) or the lower body when moving with an object while sitting (right). This informs our contact-based approach to generating object motion.}
    \label{fig:contact-weighting}
    \vspace{-1em}
\end{figure}

As visualized in Fig.~\ref{fig:contact-weighting}, object motion is naturally most influenced by parts of the human body in very close contact to the object (as they are often the cause of that motion), and less impacted (if at all) by body parts further away.
For instance, if a person moves an object with their hands, the object follows the hands but not necessarily other body parts (e.g., body and feet may remain static or walk in a different direction).
Thus, instead of directly generating one object motion hypothesis $o_i$ alongside the corresponding human motion $h_i$, we couple $o_i$ to the $M$ body contact points $j \in \{0..M-1\}$ and their predicted distances $\{c_i^j\}$ between human body surface and object geometry.

Formally, we predict object transformation hypotheses $o_i^j$ for each contact point on the human body, and weigh them with the inverse of their predicted contact distance $c_i^j$:

\begin{equation}
    o_i = \frac{1}{\sum_j \text{max}(|c_i|) - |c_i^j|} \sum_{j=0}^{M-1} (\text{max}(|c_i|) - |c_i^j|) o_i^j
\end{equation}

\subsection{Loss Formulation}
During training, the input is a noised vector $\mathbf{z}$, containing $F$ frames $\{\mathbf{x}_i\}$, each a concatenation of human body representation $h_i$, object transformation $o_i$, and contact parameters $c_i$. As condition $\mathbf{C}$, we additionally input encoded object geometry $G$ and text description $T$.
The training process is then supervised with the ground-truth sequence containing $\hat{h_i}, \hat{o_i}, \hat{c_i}$, minimizing a common objective:
\begin{equation}
    \mathbf{L} = \lambda_{h}||h_i - \hat{h_i}||_1 + \lambda_{o}||o_i - \hat{o_i}||_1 + \lambda_{c}||c_i - \hat{c_i}||_2,
\end{equation}
with $\lambda_{h}=1.0, \lambda_{o}=0.9, \lambda_{c}=0.9$.
We use classifier-free guidance \cite{ho2022classifier} for improved fidelity during inference, thus masking out the conditioning signal with $10\%$ probability.

\section{Interaction Generation}
\label{sec:interactiongen}

Using our trained network model, we can generate novel human-object interaction sequences for a given object geometry and a short text description using our weighting scheme for generating object transformations, and a custom guidance function on top of classifier-free guidance to generate physically plausible sequences.

Specifically, we use our trained model to reverse the forward diffusion process of Eq.~\ref{eq:forward-diffusion}: Starting with noised sample $\mathbf{z}_T\sim \mathcal{N}(\mathbf{0}, \mathbf{I})$, we iteratively use our trained network model $\mathcal{F}$ to estimate cleaned sample $\mathbf{z}_0$:
\begin{equation}
    \mathbf{z}_{t-1} = \sqrt{\alpha_t}\tilde{\mathbf{z}} + \sqrt{1-\alpha_t}\epsilon.
\label{eq:reverse-diffusion}
\end{equation}

\subsection{Contact-Based Diffusion Guidance}
\label{sec:guidance}
While our joint human-object-contact training already leads to plausible motions, generated sequences are not explicitly constrained to respect contact estimates during inference, which can lead to inconsistent contact between human and object motion (e.g., floating objects). 
Thus, we introduce a contact-based guidance during inference to refine predictions, using a cost function $\mathcal{G}(\mathbf{z}_t) = ||c_t-\overline{c}_t||_2^2$ which takes as input the denoised human, object, and contact predictions $\mathbf{z}_t=[h_t, o_t, c_t]$ at diffusion step $t$ and compares predicted $c_t$ and actual contact distances $\overline{c}_t$ for each contact point. Based on this, we then calculate the gradient $\nabla_{\mathbf{z}_t}\mathcal{G}(\mathbf{z}_t)$.

We use this gradient for diffusion guidance, following \cite{karunratanakul2023gmd}, by re-calculating the mean prediction $\mu_t$ at each time $t$:

\begin{equation}
\label{eq:guidance}
    \hat{\mu}_t = \mu_t + s\sum_t\nabla_{x_t}\mathcal{G}(x_t),
\end{equation}
for a scaling factor $s$. This guidance is indirect but dense in time, and is able to correct physical contact inconsistencies in the predicted sequences during inference time, without requiring any explicit post-processing steps.

\subsection{Conditioning on Object Trajectory}
\label{sec:zeroshot}

While our model has been trained with text and static object geometry as condition, we can also apply the same trained model for conditional generation of a human sequence given an object sequence and text description.  Note that this does not require any re-training, as our model has learned a strong correlation between human and object motion. 
Instead, we use a replacement-based approach, and inject the given object motion $O'$ into the diffusion process during inference at every step.
Following Eq.~\ref{eq:reverse-diffusion}, we obtain:
\begin{equation}
    \mathbf{z}_{t-1} = \sqrt{\alpha_t}\tilde{\mathbf{z}_t'} + \sqrt{1-\alpha_t}\epsilon,
\end{equation}
with $\tilde{\mathbf{z}'} = [h_t, o'_t, c_t]$, concatenating human motion $h_t$, contact distances $c_t$, and injected given object motion $o'_t$.

\section{Results}
\label{sec:experiments}
We evaluate our approach using two commonly used human-object interaction datasets BEHAVE \cite{bhatnagar2022behave} and CHAIRS \cite{jiang2022chairs} on a range of metrics, measuring motion fidelity and diversity.
We show that our approach is able to generate realistic and diverse motion on both datasets, across a variety of objects and types of interactions.

\begin{table*}[t]
\begin{center}
\resizebox{\textwidth}{!}{\begin{tabular}{|l|l||c|c|c|c||c|c|c|c|}
\hline
 & & \multicolumn{4}{c||}{BEHAVE} & \multicolumn{4}{c|}{CHAIRS} \\
\hline
\hline
Task & Approach & R-Prec. (top-3) $\uparrow$ & FID $\downarrow$ & Diversity $\rightarrow$ & MModality $\rightarrow$ & R-Prec. (top-3) $\uparrow$ & FID $\downarrow$ & Diversity $\rightarrow$ & MModality $\rightarrow$ \\
\hline\hline
& Real (human) & 0.73 & 0.09 & 4.23 & 4.55 & 0.83 & 0.01 & 7.34 & 3.00 \\
\hline
Text-Cond. & MDM~\cite{tevet2023human} & 0.52 & 4.54 & 5.44 & 5.12 & 0.72 & 5.99 & \textbf{6.83} & 3.45 \\
Human & InterDiff~\cite{xu2023interdiff} & 0.49 & 5.36 & \textbf{3.98} & 3.98 & 0.63 & 6.76 & 5.24 & 2.44 \\
Only & \textbf{Ours} & \textbf{0.60} & \textbf{4.26} & 4.92 & \textbf{4.10} & \textbf{0.78} & \textbf{5.24} & 7.90 & \textbf{3.22} \\
\hline\hline
& Real & 0.81 & 0.17 & 6.80 & 6.24 & 0.87 & 0.02 & 9.91 & 6.12 \\
\hline
Motion- & InterDiff~\cite{xu2023interdiff} & 0.68 & 3.86 & 5.62 & 5.90 & 0.67 & 4.83 & 7.49 & 4.87 \\
Cond. HOI & \textbf{Ours} & \textbf{0.71} & \textbf{3.52} & \textbf{6.89} & \textbf{6.43} & \textbf{0.79} & \textbf{4.01} & \textbf{8.42} & \textbf{6.29} \\
\hline
Text- & MDM~\cite{tevet2023human} & 0.49 & 9.21 & 6.51 & 8.19 & 0.53 & 9.23 & 6.23 & 7.44 \\
Cond. & InterDiff~\cite{xu2023interdiff} & 0.53 & 8.70 & 3.85 & 4.23 & 0.69 & 7.53 & 5.23 & 4.63 \\
HOI & \textbf{Ours} & \textbf{0.62} & \textbf{6.31} & \textbf{6.63} & \textbf{5.47} & \textbf{0.74} & \textbf{6.45} & \textbf{8.91} & \textbf{5.94} \\
\hline
\end{tabular}}
\vspace{-1em}
\caption{Quantitative comparison with state-of-the-art approaches MDM~\cite{tevet2023human} and InterDiff~\cite{xu2023interdiff}. Human Only results are evaluated only on the human pose sequence, and motion-cond. denotes predictions additionally conditioned on past observations of both human and object behavior. For metrics with $\rightarrow$, results closer to the real distribution are better. Our approach outperforms these baselines in all three settings, indicating a strong learned correlation between human and object motion.
}
\vspace{-2em}
\label{tab:baselines}
\end{center}
\end{table*}

\subsection{Experimental Setup}
\noindent
\textbf{Datasets}
We conduct our experiments on two datasets containing interactions between whole-body 3D humans and corresponding objects. CHAIRS~\cite{jiang2022chairs} captures 46 subjects as their SMPL-X \cite{smplx2019} bodies interacting with 81 different types of chairs and sofas. We extract sequences in which both human and object are in motion, yielding $\approx1300$ HOI sequences, each labeled with a text description.
We use a random 80/10/10 split along object classes, ensuring that test objects are not seen during training.
BEHAVE~\cite{bhatnagar2022behave} captures 8 participants as their SMPL-H \cite{mano2017} parameters alongside 20 different objects. This yields $\approx520$ sequences with corresponding text descriptions.
We use their original train/test split.
We sample both datasets at 20 frames per second, and generate 32 frames for CHAIRS and 64 for BEHAVE, leading to generated motion that lasts up to 3 seconds.

\smallskip
\noindent
\textbf{Implementation Details}
We train our model with batch size 64 for 600k steps ($\approx$24 hours), after which we choose the checkpoint that minimizes validation FID, following \cite{xu2023interdiff}. Our attention uses 4 heads and a latent dimension of 256. Input text is encoded using a frozen CLIP-ViT-B/32 model. For classifier-free guidance during inference time, we use a guidance scale of 2.5, which empirically provides a good trade-off between diversity and fidelity. For our inference-time contact-based guidance, we use scale $s=100.0$.

\begin{figure*}[t]
    \centering
    \includegraphics[width=0.9\textwidth]{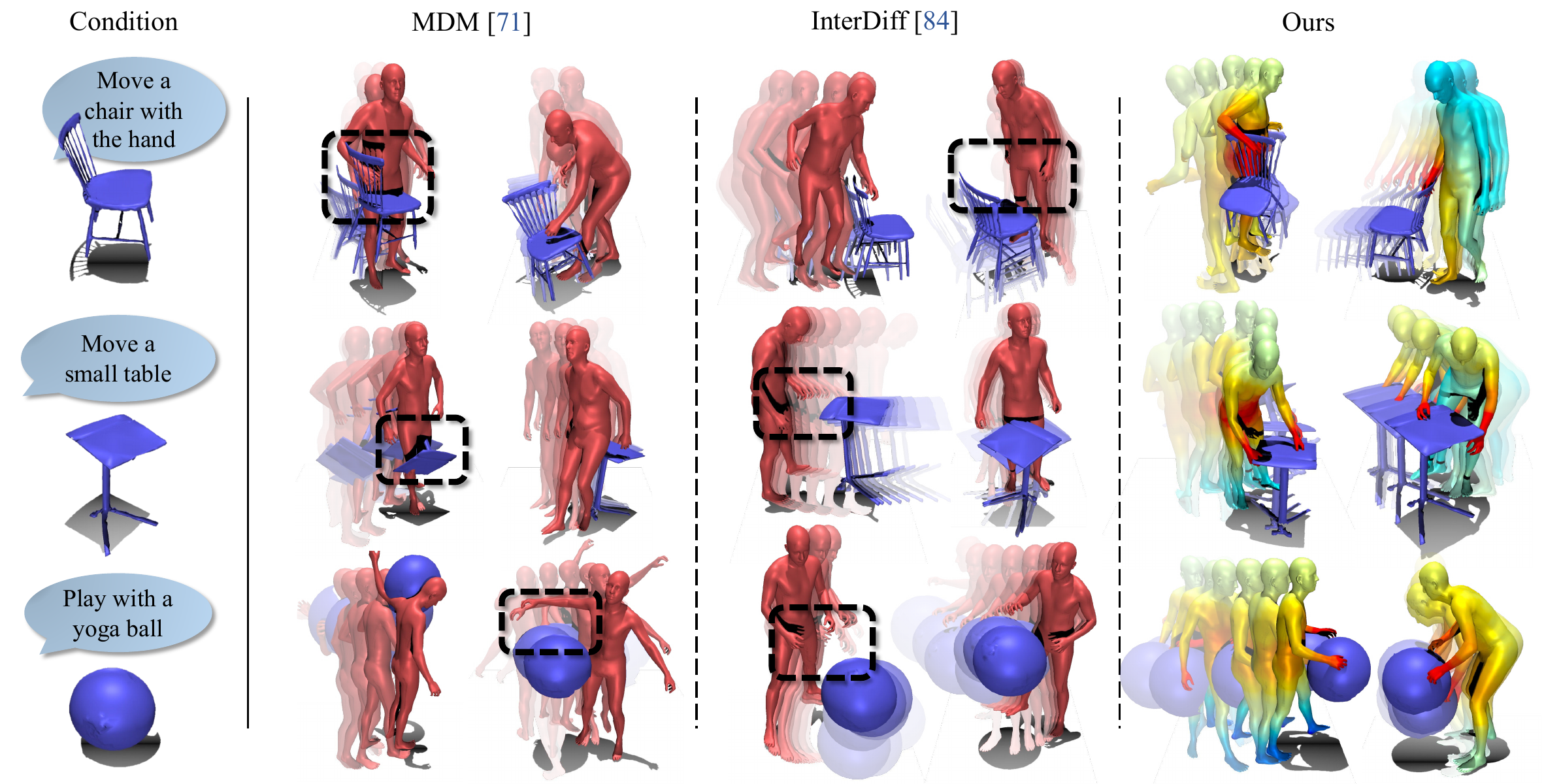}
    \vspace{-1em}
    \caption{Qualitative comparison to state-of-the-art methods MDM~\cite{tevet2023human} and InterDiff~\cite{xu2023interdiff}. Our approach generates high-quality HOIs by jointly modeling contact (closer contact in red), reducing penetration and floating artifacts (black highlight boxes).}
    \label{fig:qualitative}
    \vspace{-1em}
\end{figure*}

\subsection{Evaluation Metrics}
We measure realism and diversity of combined human and object motion, alongside closeness to the text description, following established practices~\cite{tevet2023human,guo2022generating,guo2020action2motion}. We first train a joint human-object motion feature extractor and separate text feature extractor using a contrastive loss to produce geometrically close feature vectors $\{v_i\}$ for matched text-motion pairs, and report the following metrics:

\noindent 
\textbf{R-Precision} measures the closeness of the text condition and generated HOI in latent feature space, and reports whether the correct match falls in the top 3 closest feature vectors.

\noindent
\textbf{Frechet Inception Distance (FID)}  is commonly used to evaluate the similarity between generated and ground-truth distribution in encoded feature space. 

\noindent
\textbf{Diversity and MultiModality.} Diversity measures the motion variance across all text descriptions and is defined as $\frac{1}{N}\sum_{i=1}^N||v_i-v_i'||_2$ between two randomly drawn subsets $\{v_i\}$ and $\{v_i'\}$. MultiModality (MModality) measures the average such variance intra-class, for each text description. 

\noindent
\textbf{Perceptual User Study.} The exact perceptual quality of human-object interactions is difficult to capture with any single metric; thus, we additionally conduct a user study with 32 participants to evaluate our method in comparison to baseline approaches. Participants are shown 10 baseline vs. ours pairs each in side-by side views of sequences with the same geometry and text conditioning, and asked to choose 1) Which one follows the given text better and 2) Which one looks more realistic overall. 

\begin{figure}[b]
    \centering
    \includegraphics[width=\columnwidth]{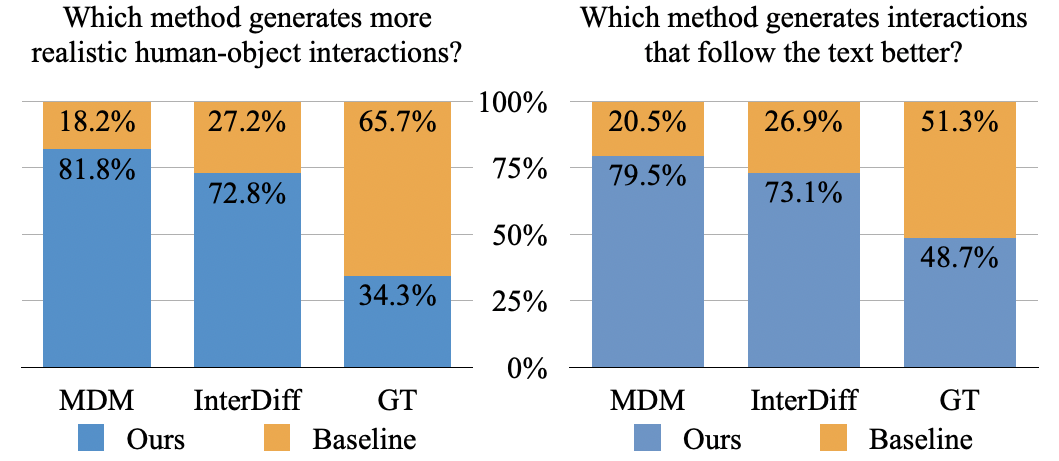}
    \vspace{-2em}
    \caption{Perceptual User Study. Participants significantly favor our method over baselines, for overall realism and text coherence.}
    \label{fig:user_study}
\end{figure}

\begin{table*}[b]
\begin{center}
\resizebox{\textwidth}{!}{\begin{tabular}{|l||c|c|c|c||c|c|c|c|}
\hline
 & \multicolumn{4}{c||}{BEHAVE} & \multicolumn{4}{c|}{CHAIRS} \\
\hline
\hline
Approach & R-Prec. (top-3) $\uparrow$ & FID $\downarrow$ & Diversity $\rightarrow$ & MModality $\rightarrow$ & R-Prec. (top-3) $\uparrow$ & FID $\downarrow$ & Diversity $\rightarrow$ & MModality $\rightarrow$ \\
\hline\hline
Real & 0.81 & 0.17 & 6.80 & 6.24 & 0.87 & 0.02 & 9.91 & 6.12 \\
\hline
No cross-attention & 0.35 & 10.44 & 8.23 & 7.40 & 0.49 & 10.84 & 12.22 & 10.64 \\
No contact prediction & 0.41 & 9.64 & 10.10 & \textbf{6.89} & 0.41 & 8.53 & 11.56 & 9.15 \\
Separate contact pred. & 0.47 & 8.01 & 5.12 & 5.12 & 0.52 & 9.34 & 7.65 & 4.62 \\
No contact weighting & 0.55 & 8.54 & 6.52 & 5.29 & 0.64 & 7.55 & 8.56 & 5.45 \\
No contact guidance & 0.59 & 7.22 & 7.84 & 5.30 & 0.70 & 7.41 & 8.05 & 5.76 \\
\hline
\textbf{Ours} & \textbf{0.62} & \textbf{6.31} & \textbf{6.63} & 5.47 & \textbf{0.74} & \textbf{6.74} & \textbf{8.91} & \textbf{5.94} \\
\hline
\end{tabular}}
\vspace{-0.3cm}
\caption{Ablation on our design choices. Joint contact prediction with cross-attention encourages the generation of more natural HOIs, and our weighting scheme and inference-time contact guidance together enable the best generation performance.}
\vspace{-1em}
\label{tab:ablations}
\end{center}
\end{table*}

\subsection{Comparison to Baselines}
As our method is the first to enable generating human and object motion from text, there are no baselines available for direct comparison. InterDiff \cite{xu2023interdiff} is closest to our approach, performing forecasting from  observed human and object motion as input and predicting a plausible continuation. In Tab.~\ref{tab:baselines}, we compare to ours first in their setting, using observed motion as condition (motion-cond.), for a fair comparison. Additionally, we modify their approach by replacing observed motion encoders with our text encoder, allowing for a comparison in our setting (text-cond.).
We also compare with MDM~\cite{tevet2023human}, a state-of-the-art method for human-only sequence generation from text, both in their original setting, only predicting human sequences, and extending theirs to also generate object sequences, by adding additional tokens and geometry conditioning to their transformer encoder formulation. For more details of baseline setup, we refer to the appendix. We evaluate the quality of generated human-object interactions as well as human-only generation, only evaluating the human sequence for our method, as compared to the generated sequences of MDM.

Both Tab.~\ref{tab:baselines} and the user study in Fig.~\ref{fig:user_study} show that our approach is able to generate more realistic and physically plausible human-object interaction sequences than baselines. In Fig.~\ref{fig:qualitative}, we see that our approach synthesizes more meaningful human-object interaction with respect to contact and mitigating independent object floating.

\subsection{Ablation Studies}

\begin{figure}[t]
    \centering
    \includegraphics[width=0.9\columnwidth]{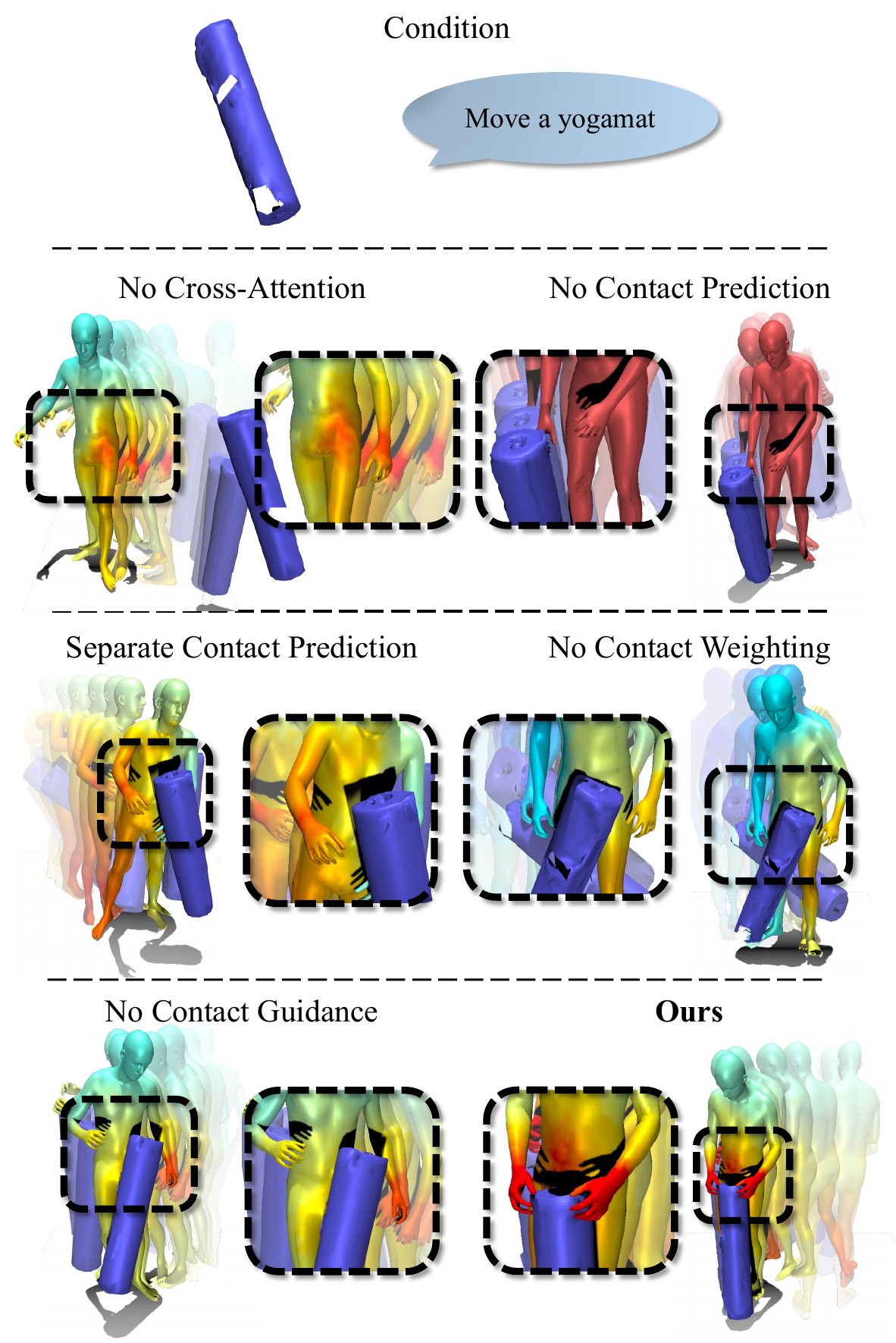}
    \caption{Visualization of ablations of our method design: Generation, weighting, and inference-time guidance work together to enable realistic interactions in our method, resolving artifacts such as object floating.}
    \vspace{-1em}
    \label{fig:ablations}
\end{figure}

\noindent
\textbf{Cross-attention enables learning human-object interdependencies.}
Tab.~\ref{tab:ablations} shows that our human-object-contact cross-attention (Sec.~\ref{sec:threeway}) significantly improves performance by effectively sharing information between human, contact, and object sequence modalities.
In Fig.~\ref{fig:ablations}, we see this encourages realistic contact between human and object.

\smallskip
\noindent
\textbf{Contact prediction improves HOI generation performance.}
Predicting contact (Sec.~\ref{sec:interactiongen}) is crucial to generating more realistic human-object sequences, resulting in more realistic interactions between human and object (Fig.~\ref{fig:ablations}), and improved fidelity (Tab.~\ref{tab:ablations}). Notably, learning contact jointly with human and object motion improves overall quality, compared to a separately trained contact model used for inference guidance (``Separate contact pred.", Tab.~\ref{tab:ablations}).

\noindent
\textbf{Contact-based object transformation weighting improves generation performance.}
Weighting predicted object motion hypotheses with predicted contact (Sec.~\ref{sec:contact-weighting}) improves HOI generation over naive object sequence prediction, both quantitatively in Tab.~\ref{tab:ablations} (``No contact weighting") and visually as realistic human-object interactions in Fig.~\ref{fig:ablations}.

\noindent
\textbf{Contact-based guidance during inference helps produce physically plausible interactions.}
As shown in Fig.~\ref{fig:ablations} and Tab.~\ref{tab:ablations}, using our guidance based on predicted contacts leads to a higher degree of fidelity and physical plausibility.

\subsection{Applications}
\begin{figure}[h]
    \centering
    \includegraphics[width=0.9\columnwidth]{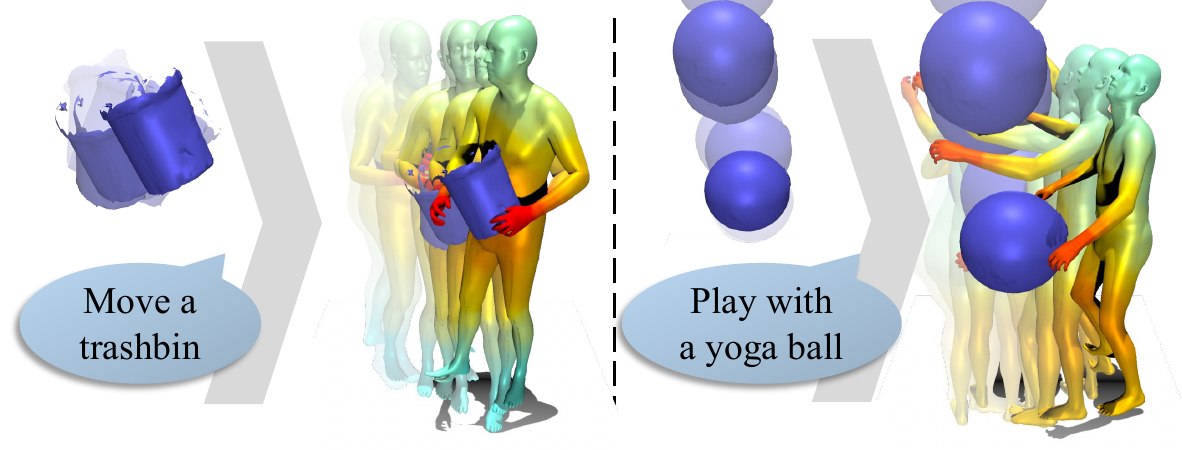}
    \caption{Given an object trajectory at inference time, our method can generate corresponding human motion without re-training.}
    \vspace{-1em}
    \label{fig:object_guidance}
\end{figure}

\noindent
\textbf{Human motion generation given object trajectory.}
Our approach can be directly applied to conditionally generate human sequences given object sequences as condition, as shown in Fig.~\ref{fig:object_guidance}. As our model learns a strong correspondence between object and human motion, facilitated by contact distance predictions, we are able to condition without any additional training.

\begin{figure}[h]
    \centering
    \includegraphics[width=0.9\columnwidth]{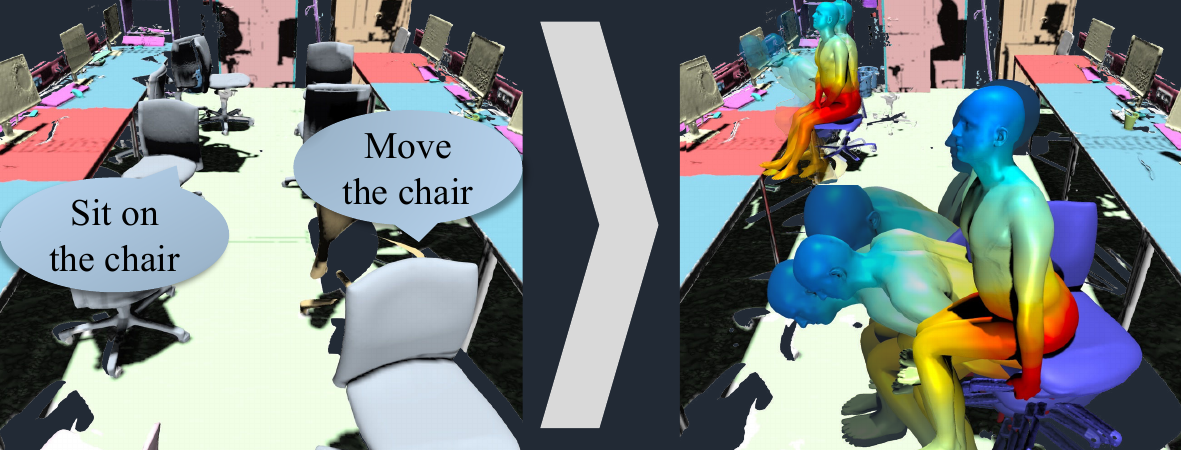}
    \caption{Application to static 3D scene scans. Our method can generate HOIs from segmented objects in such environments.}
    \vspace{-1em}
    \label{fig:scene_population}
\end{figure}

\noindent
\textbf{Populating 3D scans.} 
Fig.~\ref{fig:scene_population} shows that we can also apply our method to generate human-object interactions in static scene scans. Here, we use a scene from the ScanNet++ dataset~\cite{yeshwanth2023scannetpp}, with their existing semantic object segmentation.
This enables the potential to generate realistic human motion sequences only given a static scene environment.

\subsection{Limitations}
\label{limitations}

While we have demonstrated the usefulness of joint contact prediction in 3D HOI generation, several limitations remain. For instance, our method focuses on realistic interactions with a single object. We show that this can be applied to objects in static 3D scans; however, we do not model multiple objects together, which could have the potential to model more complex long-term human behavior (e.g. cooking sequences). Additionally, our method requires expensive 3D HOI captures for training; a weakly supervised approach leveraging further supervision from 2D action data might be able to represent more diverse scenarios. Similarly, our method depends on manual text annotations; more specific prompts might lead to more control over generated HOIs.

\section{Conclusion}
\label{sec:conclusion}

We propose an approach to generating realistic, dynamic human-object interactions based on contact modeling. Our diffusion model effectively learns interdependencies between human, object, and contact through cross-attention along with our contact-based object transformation weighting.
Our predicted contacts further facilitate refinement using custom diffusion guidance, generating diverse, realistic interactions based on text descriptions.
Since our model learns a strong correlation between human and object sequences, we can use it to conditionally generate human motion from given object sequences.
Extensive experimental evaluation confirms both fidelity and diversity of our generated sequences and shows improved performance compared to baselines.

\section*{Acknowledgements}
\label{sec:acknowledgements}

This project is funded by the ERC Starting Grant SpatialSem (101076253), the Bavarian State Ministry of Science and the Arts coordinated by the Bavarian Research Institute for Digital Transformation (bidt), and the German Research Foundation (DFG) Grant ``Learning How to Interact with Scenes through Part-Based Understanding".

{
    \small
    \bibliographystyle{ieee_fullname}
    \bibliography{main}
}

\clearpage
\renewcommand{\appendixpagename}{\Large{Appendix}} 

\begin{appendices}
We show in this appendix additional qualitative (Sec.~\ref{sec:additional-qualitative-results}) and quantitative (Sec.~\ref{sec:additional-quantitative-results}) results, detail our baseline evaluation protocol (Sec.~\ref{sec:baseline-evaluation}), elaborate on the metrics used in the main paper (Sec.~\ref{sec:metric-details}), show the architecture used in our approach (Sec.~\ref{sec:architecture}), and provide additional details regarding the data (Sec.~\ref{sec:data-details}).
\section{Additional Qualitative Results}
\label{sec:additional-qualitative-results}
\subsection{Additional Interactions}
\begin{figure*}
    \centering
    \includegraphics[width=0.8\textwidth]{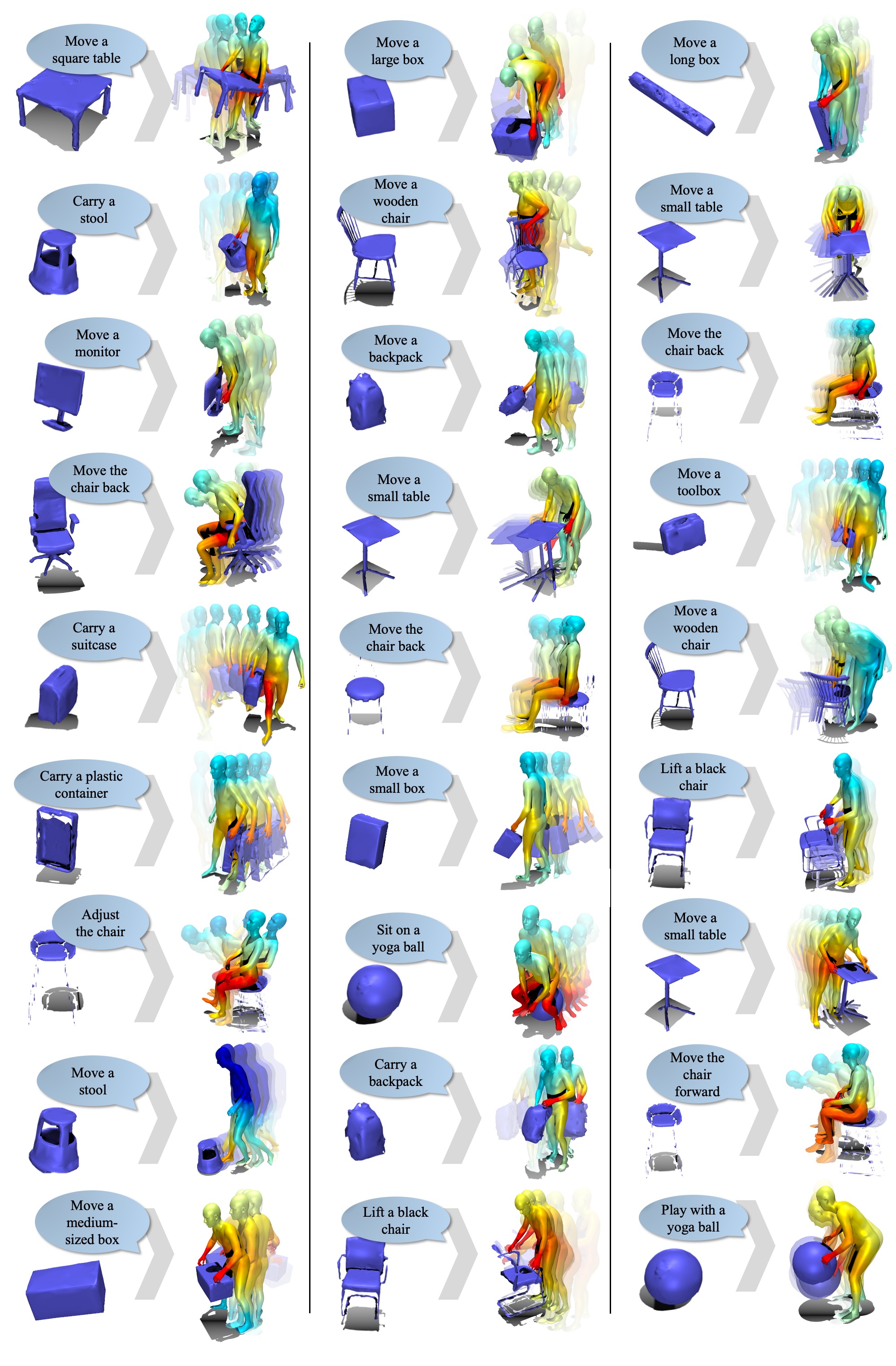}
    \vspace{-1em}
    \caption{Additional qualitative evaluation. Our method produces diverse and realistic 3D human-object interaction sequences, given object geometry and short text description of the action. The sequences depict high-quality human-object interactions by modeling contact, mitigating floating and penetration artifacts.}
    \label{fig:additional-qualitative}
\end{figure*}

We show additional generated 3D human-object interactions of our method in Fig.~\ref{fig:additional-qualitative}, with object geometry and text condition on the left, and our generated sequence on the right.

\subsection{Same Prompt, Different Interactions}
\begin{figure}[h]
    \centering
    \includegraphics[width=\columnwidth]{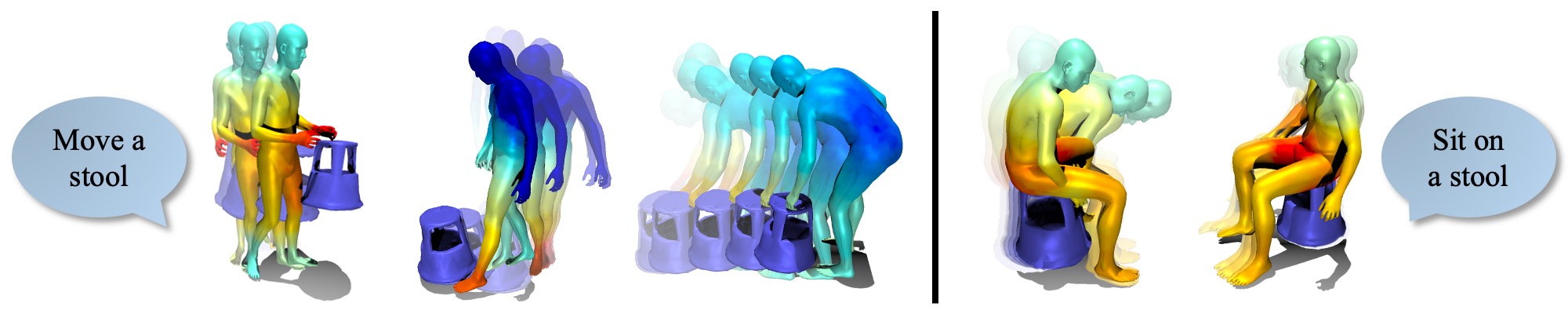}
    \caption{Our method is able to generate diverse human-object interactions for the same prompts.}
    \label{fig:move-stool}
    \vspace{-1em}
\end{figure}

We evaluate the ability of our method to generate diverse interactions for a fixed text condition visually in Fig.~\ref{fig:move-stool}, for text prompt ``Move a stool'' and ``Sit on a stool''. In the ground truth training data, move is only done with one or two hands, and feet; moving with the butt sometimes occurs for the text description ``Sit on a stool''.

\section{Additional Quantitative Results}
\label{sec:additional-quantitative-results}
\subsection{Evaluating Penetrations and Floating}
Our method discourages penetration and floating implicitly, by enforcing correct contact distances as a soft constraint at train and test time. However, the exact fidelity and diversity of our results is hard to capture with any single metric. Thus, we evaluate multiple such metrics in the main paper (R-Precision, FID, Diversity, MultiModality), and conduct a perceptual user study to verify the metrics' expressiveness.

Here, we provide an additional evaluation based on intuitive physics-based metrics: Tab.~\ref{tab:penetration-evaluation} evaluates the mean ratio of frames with some penetration as well as the ratio of penetrating vertices overall, showing that penetrations typically happens with small body parts (e.g., hands, which also occurs in the ground-truth data). 
We also evaluate the ratio of frames and vertices with human and object not in contact, including floating and stationary objects, which is expected to be close to the dataset ratio.

Results show similar penetration and floating between our generations and ground-truth training data.

\begin{table}[h]
\begin{center}
\resizebox{\columnwidth}{!}{\begin{tabular}{|l||c|c|c|c||c|c|c|c|}
\hline
 & \multicolumn{4}{c||}{BEHAVE} & \multicolumn{4}{c|}{CHAIRS} \\
\hline
\hline
  & \multicolumn{2}{c|}{Penetration} & \multicolumn{2}{c||}{Non-Contact} & \multicolumn{2}{c|}{Penetration} & \multicolumn{2}{c|}{Non-Contact} \\
\hline
 & Frames & Vertices & Frames & Vertices & Frames & Vertices & Frames & Vertices \\
\hline
Dataset & 32.9\% & 4.1\% & 21.4\% & 86.2\% & 26.9\% & 1.1\% & 11.9\% & 70.4\% \\
\textbf{Ours} & 31.3\% & 3.0\% & 17.8\% & 93.3\% & 35.8\% & 4.2\% & 14.1\% & 74.3\% \\
\hline
\end{tabular}}
\caption{
Penetration and non-contact (including floating) ratios in terms of frames as well as overall vertices vs ground-truth data.
}
\vspace{-2em}
\label{tab:penetration-evaluation}
\end{center}
\end{table}

\subsection{Evaluating Contact}
Tab.~\ref{tab:contact-evaluation} evaluates our contact predictions using precision/recall and distance metrics.
We follow \cite{hassan2021populating,zhang2020place,zhang2020generating} to define contact if~${\le}5$cm from object. We also report mean $\ell_1$ error in contact distance predictions. All metrics are reported for body parts ${\leq}1m$ of the object, to focus on contact scenarios. Better contact prediction corresponds with better HOI generations.
Note that none of our baselines predict contact distances.

\begin{table}[h]
\begin{center}
\resizebox{\columnwidth}{!}{\begin{tabular}{|l||c|c|c||c|c|c|}
\hline
 & \multicolumn{3}{c||}{BEHAVE} & \multicolumn{3}{c|}{CHAIRS} \\
\hline
\hline
Approach & Precision $\uparrow$ & Recall $\uparrow$ & Distance $\downarrow$ & Precision $\uparrow$ & Recall $\uparrow$ & Distance $\downarrow$ \\
\hline
Separate contact pred. & 23.4\% & 25.6\% & 0.53 & 58.6\% & 49.1\% & 0.24 \\
No contact weighting & 29.5\% & 33.5\% & 0.34 & 60.6\% & 63.4\% & 0.10 \\
No contact guidance & 46.3\% & 39.2\% & 0.31 & 64.2\% & 70.2\% & 0.12 \\
\hline
\textbf{Ours} & \textbf{63.6\%} & \textbf{59.5\%} & \textbf{0.07} & \textbf{78.3\%} & \textbf{84.5\%} & \textbf{0.04} \\
\hline
\end{tabular}}
\caption{
Evaluation of predicted contact distances, in terms on precision and recall ($\le 5cm$ distance), as well as mean contact $\ell_1$ error in meters.
}
\vspace{-2em}
\label{tab:contact-evaluation}
\end{center}
\end{table}

\subsection{Novelty of Generated Interactions}
We perform an additional interaction novelty analysis to verify that our method does not simply retrieve memorized train sequences but is indeed able to generate novel human-object interactions. To do so, we generate $\approx500$ sequences from both datasets and retrieve the top-3 most similar train sequences, as measured by the $l_2$ distance in human body and object transformation parameter space.

\begin{figure*}
    \centering
    \includegraphics[width=\textwidth]{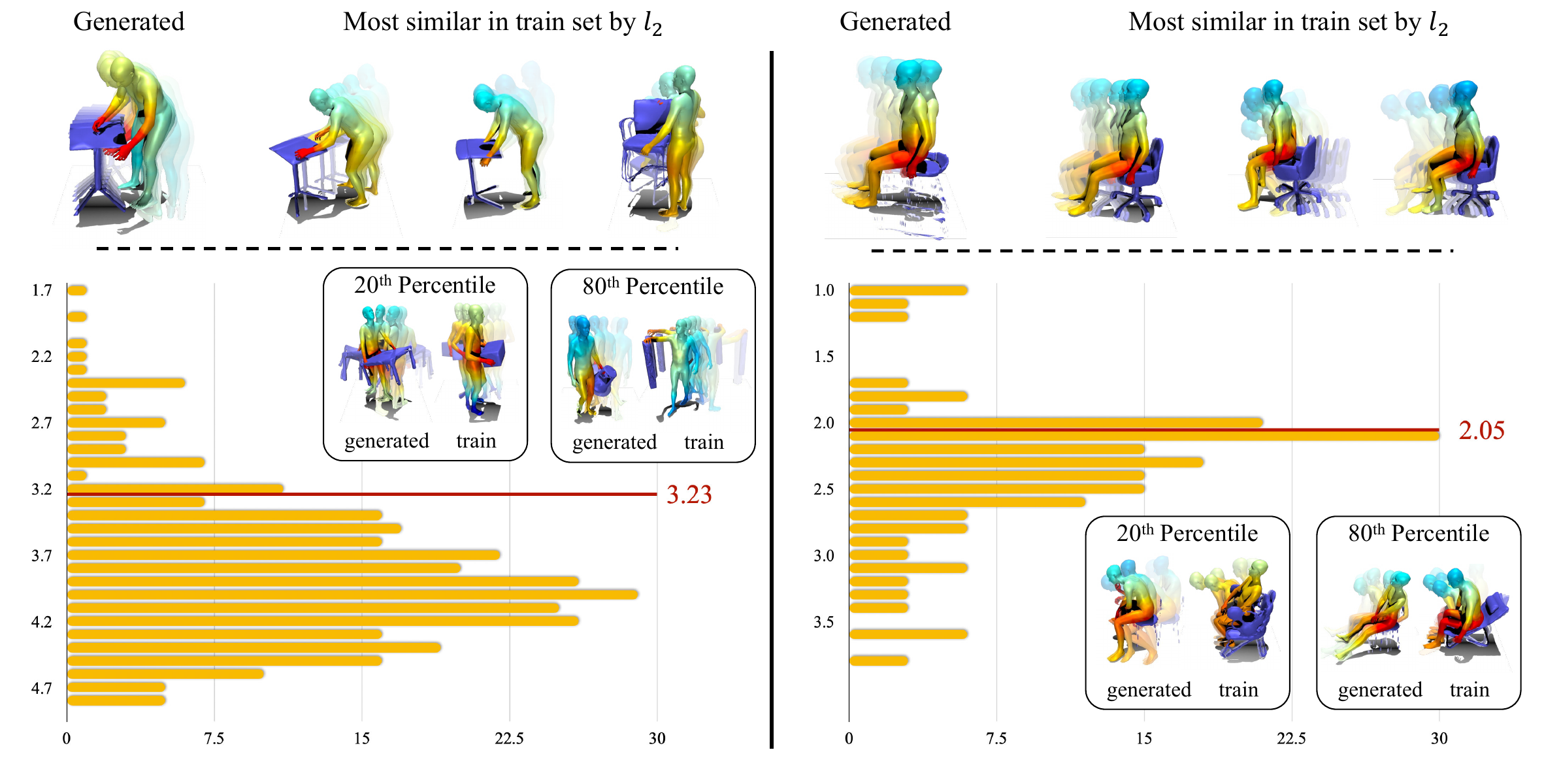}
    \vspace{-1em}
    \caption{Human-Object Interaction Sequence Novelty Analysis. Performed on BEHAVE~\cite{bhatnagar2022behave} (left) and CHAIRS~\cite{jiang2022chairs} (right). We retrieve top-3 most similar sequences from the train set, and plot a histogram of distances to the closest train sample. While sequences at the 20th percentile still resemble the generated interactions, there is a large gap in the 80th percentile. We show the intra-trainset distance in red. Our approach generates novel shapes, not simply retrieving memorized train samples.}
    \label{fig:retrieval}
\end{figure*}

Fig.~\ref{fig:retrieval} shows the top-3 closest train sequences, along with a histogram of $l_2$ distances computed on our test set of $\approx500$ generated sequences. In red, we mark the intra-trainset distance between samples in the train set.
We observe that the distance between our generated sequences and the closest train sequence is mostly larger than the intra-train distance. Thus, our method is able to produce samples that are novel and not simply retrieved train sequences.

\subsection{SMPL Bodies vs. HumanML3D Skeletons}
We observe slight pose jitter and foot skating in our ground-truth training data (especially BEHAVE, captured with Kinect sensors). As a result, our model reflects some of these effects. 
Skeleton representations such as HumanML3D~\cite{guo2022generating} could tackle these artifacts, but do not work with contact as effectively as SMPL bodies. Nevertheless, we train ours with HumanML3D parameters for comparison in Tab.~\ref{tab:humanml3d} (fitting SMPL after for comparable evaluation) which leads to degraded performance due to less effective contact guidance.

\begin{table*}[b]
\begin{center}
\resizebox{0.9\textwidth}{!}{\begin{tabular}{|l||c|c|c|c||c|c|c|c|}
\hline
 & \multicolumn{4}{c||}{BEHAVE} & \multicolumn{4}{c|}{CHAIRS} \\
\hline
\hline
Representation & R-Prec. (top-3) $\uparrow$ & FID $\downarrow$ & Diversity $\rightarrow$ & MModality $\rightarrow$ & R-Prec. (top-3) $\uparrow$ & FID $\downarrow$ & Diversity $\rightarrow$ & MModality $\rightarrow$ \\
\hline
Ours (HumanML3D) & 0.33 & 11.94 & 2.15 & 3.75 & 0.48 & 12.83 & 4.39 & 5.11 \\
\textbf{Ours} & \textbf{0.62} & \textbf{6.31} & \textbf{6.63} & \textbf{5.47} & \textbf{0.74} & \textbf{6.45} & \textbf{8.91} & \textbf{5.94} \\
\hline
\end{tabular}}
\caption{
Ours (using SMPL bodies) vs. using HumanML3D~\cite{guo2022generating} skeletons and fitting SMPL bodies afterwards. While HumanML3D is designed to reduce jitter and foot skating, it leads to degraded performance in our scenario due to less effective contact guidance.
}
\label{tab:humanml3d}
\vspace{-2.25em}
\end{center}
\end{table*}

\section{Baseline Evaluation Setup}
\label{sec:baseline-evaluation}
There is no previous approach to modeling 3D human-object interactions from text and object geometry for direct comparison. Thus, we compare to the two closest methods, and compare to them in multiple settings, for a fair comparison.

The most related approach is InterDiff~\cite{xu2023interdiff}. Their setting is to generate a short sequence of human-object interactions, from an observed such sequence as condition, with geometry but no text input. Their goal is to generate one, the most likely, sequence continuing the observation. We use their full approach, including the main diffusion training together with the post-processing refinement step.
We compare in two different settings: First, in their native setup, running their method unchanged and modifying ours to take in geometry and past sequence observation instead of text (Motion-Cond. HOI in Tab.~1 main). 
Then, we modify their approach to take in geometry and text, replacing their past motion encoder with our CLIP-based text encoder (Text-Cond. HOI in Tab.~1 main). We observe that our method is able to outperform InterDiff in both scenarios, for both datasets.

We additionally compare to MDM~\cite{tevet2023human}, a recent diffusion-based state-of-the-art human motion generation approach. Their approach is based on a transformer encoder formulation, using each human body as a token in the attention. We run their method on SMPL parameters and first compare in their native setting, only predicting human motion. We compare to the human motion generated by our method which is trained to generate full human-object interactions (Text-Cond. Human Only in Tab.~1 main). We also compare to human motion sequences generated by InterDiff in this setting. We see that our method is able to outperform both baselines even in this setting, demonstrating the added benefit of learning interdependencies of human and object motion. 
For the comparison in our setting, we modify MDM by adding additional tokens for the objects to the attention formulation. Our approach performs more realistic and diverse sequences in both settings which better follow the text condition.

\begin{figure*}[b]
    \centering
    \includegraphics[width=\textwidth]{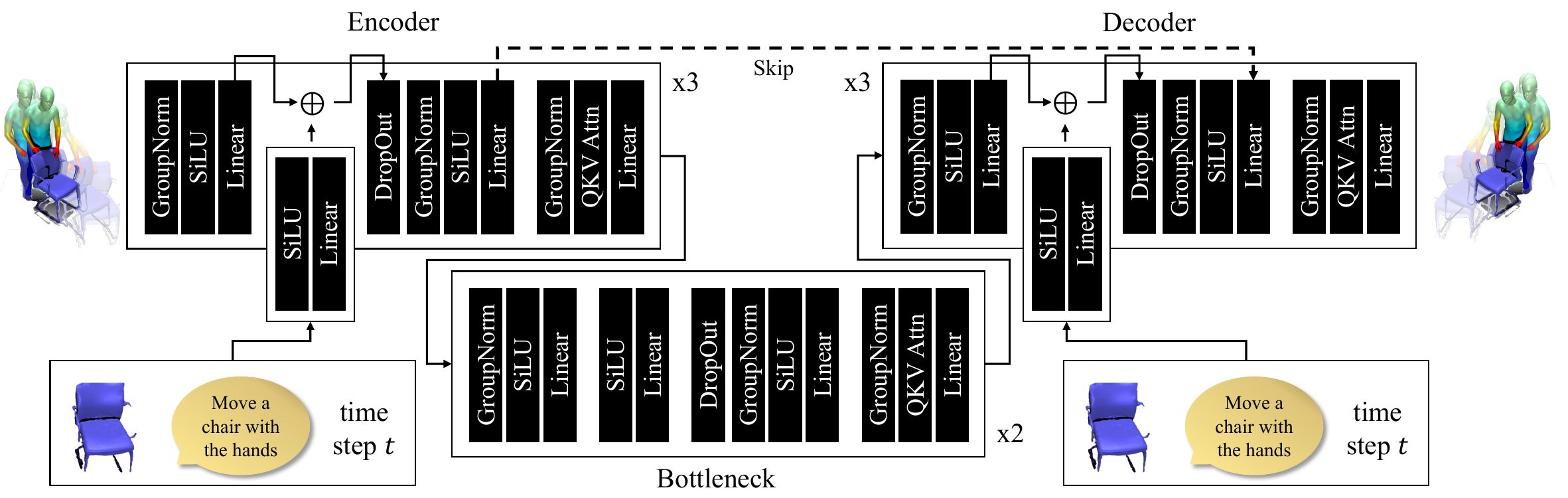}
    \vspace{-1em}
    \caption{Network architecture specification.}
    \label{fig:architecture}
\end{figure*}

\section{Fidelity and Diversity Metrics}
\label{sec:metric-details}
We base our fidelity and diversity metrics R-Precision, FID score, Diversity, and MultiModality on practices established for human motion generation~\cite{tevet2023human,guo2022generating,guo2020action2motion}, with minor modifications: We use the same networks used by these previous approaches, and adapt the input dimensions to fit our feature lengths, $F=79$ when evaluating human body motion only, and $F=79+128+9=216$ (SMPL parameters, contact distances, object transformations) for full evaluation in the human-object interaction scenario.

\section{Architecture Details}
\label{sec:architecture}
Fig.~\ref{fig:architecture} shows our detailed network architecture, including encoder, bottleneck, and decoder formulations.

\section{Data Details}
\label{sec:data-details}
\subsection{Datasets}
\paragraph{CHAIRS~\cite{jiang2022chairs}}
captures 46 subjects as their SMPL-X \cite{smplx2019} parameters using a mocap suit, in various settings interacting with a total of 81 different types of chairs and sofas, from office chairs over simple wooden chairs to more complex models like suspended seating structures. Each captured sequence consists of 6 actions and a given script; the exact separation into corresponding textual descriptions was manually annotated by the authors of this paper. 
In total, this yields $\approx1300$ sequences of human and object motion, together with a textual description.
Every object geometry is provided as their canonical mesh; we additionally generate ground-truth contact and distance labels based on posed human and object meshes per-frame for each sequence.
We use a random 80/10/10 split along object types, making sure that test objects are not seen during training.

\paragraph{BEHAVE~\cite{bhatnagar2022behave}}
captures 8 participants as their SMPL-H \cite{mano2017} parameters captured in a multi-Kinect setup, along with the per-frame transformations and canonical geometries of 20 different object with a wide range, including yoga mats and tables.
This yields $\approx130$ longer sequences.
We use their original train/test split.

\subsection{Object Geometry Representation}
We represent object geometry as a point cloud, to be processed by a PointNet~\cite{pointnet2017} encoder. For this, we sample $N=256$ points uniformly at random on the surface of an object mesh. Each object category is sampled once as a pre-processing step and kept same for train and inference.

\end{appendices}

\end{document}